\documentclass[runningheads]{llncs}
\usepackage[T1]{fontenc}
\usepackage{graphicx}
\usepackage{hyperref}
\usepackage{xcolor}
\usepackage{soul}
\usepackage{caption}
\usepackage{subcaption}

\definecolor{colorMatej}{RGB}{100, 5, 168}

\begin{document}
\title{
Challenges in Explaining Pretrained Clinical Text Classifiers
}
%
%
\author{Kristian Miok\inst{1,2} 
\and
Matej Klemen\inst{1} 
\and
Blaz \v Skrlj \inst{3} 
\and
Marko Robnik \v Sikonja\inst{1}
}
%
%
\institute{Faculty of Computer and Information Science, University of Ljubljana, Slovenia\\
\email{\{kristian.miok,matej.klemen,marko.robnik\}@fri.uni-lj.si}
\and
ICAM - Advanced Environmental Research Institute, West University of Timisoara, Romania 
\email{kristian.miok@e-uvt.ro}\\
\and
Jo\v zef Stefan Institute \\
\email{blaz.skrlj@ijs.si}}

\maketitle              
\renewcommand{\thefootnote}{}
\footnotetext{To appear in the Proceedings of the First Workshop on Responsible Healthcare using Machine Learning (RHCML 2025)}

\begin{abstract}
Explaining the predictions of neural models in clinical NLP remains a significant challenge, especially for complex tasks involving long, unstructured medical texts. While post-hoc methods like LIME and SHAP are widely used, they often fall short when applied to clinical narratives. In this paper, we identify core limitations of token-level and perturbation-based explanation techniques through targeted demonstrations on a hospital length-of-stay prediction task. Our findings reveal issues such as overemphasis on non-informative tokens, instability in attributions, and high-confidence predictions for incoherent input variants. These results underscore the need for explanation strategies that are clinically meaningful, semantically grounded, and robust to linguistic noise.
\keywords{Explainable AI \and Clinical NLP \and LIME \and Electronic Health Records}
\end{abstract}

\section{Introduction}

Deep learning models achieve strong predictive performance, yet their use in high-stakes domains such as healthcare remains limited without clear explanations of their decisions. Explainable AI (XAI) seeks to increase transparency by revealing how predictions are made, enabling reliability checks, bias detection, and accountability \cite{bhattacharya2022applied}. In medicine, where decisions affect patient outcomes, interpretable models are particularly critical.

Text-based electronic health records (EHRs) provide valuable clinical information, including notes, discharge summaries, and pathology reports \cite{tayefi2021challenges,benedum2023replication}. Automated analysis of these documents supports tasks such as incident classification \cite{palojoki2021classification} and long-document modeling with transformers \cite{dai2022revisiting}. Despite such progress, the explainability of clinical text classifiers remains underexplored. Most XAI techniques were designed for tabular or image data, and when applied to long, noisy narratives, token-level approaches like LIME and SHAP often highlight trivial tokens, ignore clinical concepts, or produce unstable results.

In this paper, we examine the limitations of post-hoc explanations in clinical NLP, focusing on hospital discharge summaries. Section \hyperref[General]{2} outlines key challenges, Section \hyperref[Demonstration]{3} demonstrates them on a length-of-stay prediction task, and Section \hyperref[conclusion]{4} concludes with broader implications for responsible ML in healthcare.

\section{Related Work}

Post-hoc explanation methods such as LIME \cite{ribeiro2016lime} and SHAP \cite{lundberg2017shap} are widely used to attribute importance to input features, but their application to text often inherits limitations from token-level representations. In clinical NLP, prior studies have applied these tools to tasks such as diagnosis prediction and patient risk stratification \cite{caruana2015intelligible,tonekaboni2019clinician}, yet results frequently highlight stopwords or isolated tokens rather than clinically meaningful units. More recent efforts have explored phrase-level or concept-based explanations, e.g., by aggregating token scores into medical entities or leveraging ontologies \cite{holzinger2022knowledge}, which better align with expert reasoning. Nonetheless, explainability for long and heterogeneous clinical narratives remains underexplored compared to domains such as tabular or imaging data, underscoring the need for approaches that are linguistically coherent, semantically grounded, and clinically valid.

\section{Challenges}
\label{General}

Popular explanation methods such as LIME \cite{ribeiro2016lime} and SHAP \cite{lundberg2017shap} assign importance scores to input features. While effective for tabular data, their direct application to text, especially long and noisy clinical narratives, introduces several limitations:

\begin{enumerate}
    \item \textbf{Interpretation of feature attributions.} Only a few top-ranked tokens meaningfully affect predictions, while many others with non-negligible scores contribute little. This uneven relevance risks over-interpreting weak or ambiguous features.
    
    \item \textbf{Length and complexity.} Clinical reports contain shorthand, numbers, and structured fragments. Token-level explanations often highlight trivial terms, limiting their clinical utility.
    
    \item \textbf{Longer textual units.} Many medical concepts appear as multi-word expressions (e.g., “chronic kidney disease”). Token-based attribution fails to capture such phrases, instead emphasizing isolated or function words.
    
    \item \textbf{Off-manifold perturbations.} Perturbation strategies can create incoherent or grammatically broken text that still yields confident predictions, distorting the explanation process.
    
    \item \textbf{Dilution through averaging.} Averaging effects across many perturbed samples can obscure important localized patterns, producing diffuse or inconsistent explanations.
    
    \item \textbf{Instability.} Because LIME relies on random sampling, repeated runs may yield different attributions, undermining reproducibility in clinical settings.
\end{enumerate}

\section{Demonstration of Challenges}
\label{Demonstration}

To better understand the behavior and limitations of post-hoc explanation methods in clinical NLP, we conduct a series of targeted demonstrations using a real-world dataset of hospital discharge summaries. Our focus is on the task of predicting hospital length of stay (LOS) based solely on free-text notes. The dataset includes 467 de-identified discharge summaries from adult patients diagnosed with kidney stones (ICD-10 code N20.0), sourced from the MIMIC-IV database. Each summary is labeled according to whether the patient's hospital stay exceeded the cohort median, forming a balanced binary classification task.

This dataset reflects many typical complexities of clinical text: highly variable document lengths (ranging from hundreds to over a thousand tokens), domain-specific terminology, shorthand notations, embedded numerical values, and a mixture of structured and narrative content. These properties make it a representative and challenging testbed for evaluating explanation methods.

We fine-tuned a domain-adapted \texttt{ModernBERT} model on this classification task using the HuggingFace Transformers framework. The model was trained on full-length discharge summaries truncated to 1024 tokens and optimized using cross-entropy loss. To interpret the model’s predictions, we employed LIME (Local Interpretable Model-Agnostic Explanations), which constructs local surrogate models by perturbing input tokens and fitting a sparse linear classifier.

In this section, we present three demonstrations that empirically expose recurring issues in how LIME explanations behave on clinical text:

\begin{itemize}
    \item \textbf{Demonstration 1}: We assess how model confidence changes when top-ranked LIME tokens are progressively removed, revealing that only a small number of tokens drive predictions while the rest contribute minimally.
    
    \item \textbf{Demonstration 2}: We analyze the frequency of tokens selected by LIME across multiple examples and show that stopwords and function words often dominate explanations, despite offering limited interpretive value.
    
    \item \textbf{Demonstration 3}: We generate perturbations of a synthetic but medically plausible note and find that semantically incoherent or fragmented inputs can still yield high-confidence predictions, raising concerns about explanation reliability.
\end{itemize}

Each demonstration is supported by quantitative metrics and qualitative visualizations to reveal patterns in model behavior and explainer outputs. Collectively, they illustrate systematic weaknesses in current explanation approaches and motivate the need for methods that are more aligned with clinical reasoning and human interpretability.

\subsubsection{Demonstration 1: Sensitivity of model confidence to top-ranked tokens}
\label{dem1}


To better understand how LIME’s token-level explanations correspond to model behavior, we conducted a deletion-based sensitivity analysis. After fine-tuning a \texttt{ModernBERT} model to classify hospital discharge summaries as predicting either a short or long length of stay, we selected 20 representative test instances and generated LIME explanations for each.

For each case, we extracted the top-10 most influential tokens as identified by LIME and progressively removed them from the input, one by one, based on their importance rank. At each step, we recorded the model's prediction confidence for the originally predicted class. This yielded a deletion curve representing how quickly the model's confidence degraded as influential features were masked. For comparison, we also constructed deletion curves using randomly selected tokens.

\begin{figure}[t]
\centering
\includegraphics[width=0.65\textwidth]{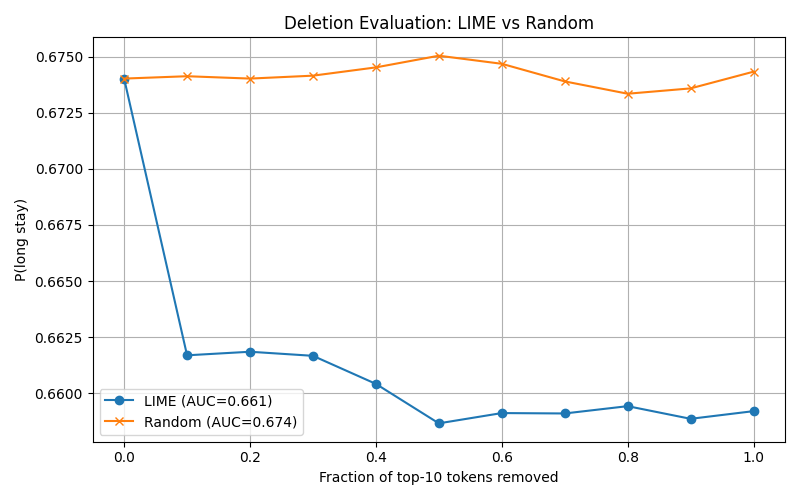}  
\caption{Deletion curves comparing LIME-guided and random token removal for 20 test instances. Removing the top-ranked LIME tokens causes a sharp initial drop in predicted probability, whereas randomly selected tokens have a weaker and more uniform effect.}
\label{fig:cha1}
\end{figure}

Figure~\ref{fig:cha1} illustrates the outcome. We observed that removing the top one or two tokens typically caused a sharp decline in prediction confidence, suggesting these few tokens were strongly driving the decision. Beyond the top three or four tokens, however, the effect largely plateaued, with additional removals leading to only marginal changes. Meanwhile, deletion of random tokens led to a gentler and more gradual decline.

This finding suggests that while a small number of tokens meaningfully influence the model’s output, the remaining high-attribution tokens contribute little explanatory value. In contexts where users might examine the entire ranked list of important tokens, this effect may lead to overinterpretation of marginal or irrelevant features. Effective explanation strategies should help users distinguish between dominant drivers of the model’s output and tokens with negligible influence.

\subsubsection{Demonstration 2: Prevalence of non-informative tokens in explanations}
\label{dem2}

We examined the interpretability of LIME-generated explanations by analyzing which tokens received the highest attribution scores across a sample of 20 clinical discharge summaries. From each explanation, we extracted the top-10 most influential tokens, yielding 200 tokens in total.

\begin{figure}[t]
\centering
\begin{subfigure}{0.48\textwidth}
\includegraphics[width=\textwidth]{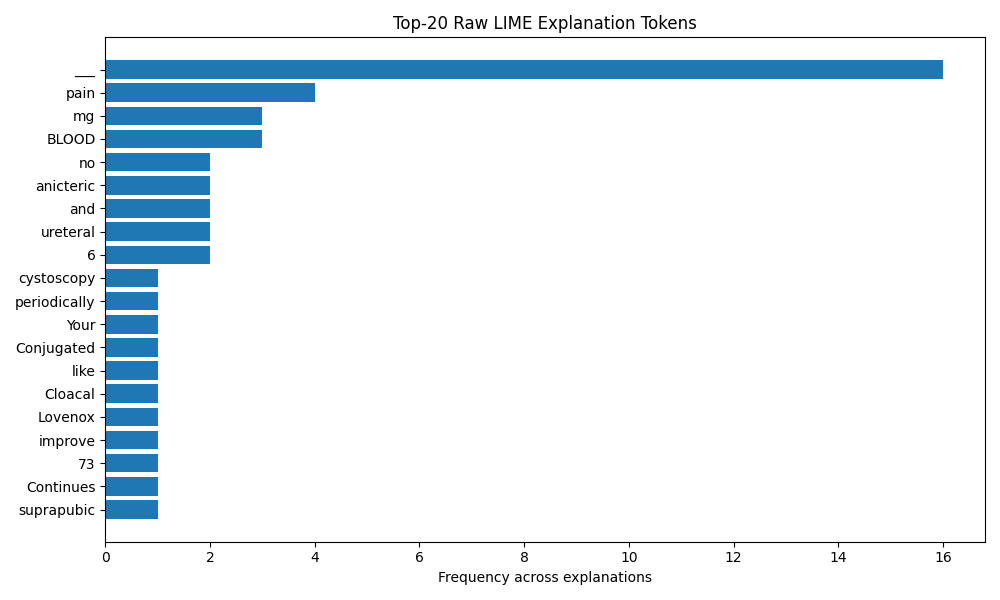}
\caption{Top-20 LIME tokens without filtering (includes stopwords and symbols).}
\label{fig:token-barchart-raw}
\end{subfigure}
\hfill
\begin{subfigure}{0.48\textwidth}
\includegraphics[width=\textwidth]{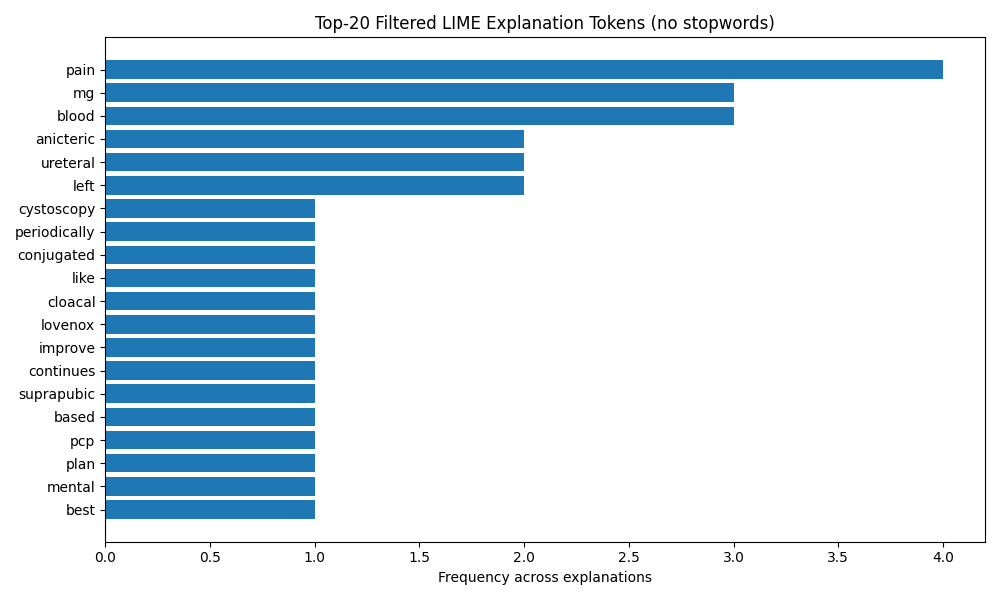}
\caption{Top-20 LIME tokens after removing stopwords and symbols.}
\label{fig:token-barchart-filtered}
\end{subfigure}
\caption{Distribution of common LIME attribution tokens across 20 clinical discharge summaries. The filtered view better highlights meaningful medical terms.}
\label{fig:token-barplot-comparison}
\end{figure}

We aggregated these tokens to identify frequently highlighted words and visualized the results in two complementary plots:

\begin{itemize}
\item The \textbf{raw bar chart} in Figure~\ref{fig:token-barchart-raw} shows the 20 most frequent LIME-selected tokens before any preprocessing. Several tokens are non-informative, including formatting artifacts (e.g., underscores), numbers, and function words (e.g., \texttt{no}, \texttt{and}, \texttt{Your}), which offer limited clinical interpretability.

\item The \textbf{filtered bar chart} in Figure~\ref{fig:token-barchart-filtered} presents the same analysis after removing stopwords, non-alphabetic tokens, and trivial words. The resulting top tokens include medically meaningful terms such as \texttt{pain}, \texttt{blood}, \texttt{cystoscopy}, and \texttt{ureteral}, which are more plausible indicators of clinical reasoning behind model predictions.
\end{itemize}

This demonstration illustrates a key weakness of token-level explanation strategies: non-informative tokens often receive high attribution simply due to their frequency or structural placement, not semantic relevance. Although filtering can improve clarity, it also risks discarding potentially meaningful expressions that only gain interpretability in context (e.g., as part of multi-word medical phrases). Therefore, reliance on isolated token attributions can lead to explanations that are misleading or unhelpful in practice. Modern NLP methods can detect multi-word expressions or medical entities before stopword removal, which may provide a more faithful basis for generating clinically meaningful explanations.

\subsubsection{Demonstration 3: Nonsensical permutations with high-confidence predictions}

\label{dem3}

To further examine the effects of off-manifold perturbations, we created a synthetic but medically coherent discharge summary for a hypothetical patient. This artificial note reflects typical clinical phrasing and content, but allows us to observe LIME’s behavior under controlled conditions. The original note reads as follows:

\begin{quote}
\texttt{Patient is a 65-year-old female with history of type 2 diabetes, hypertension, and chronic kidney disease. She presented with chest pain and dyspnea. Vitals stable. Labs: elevated troponin and CK-MB. Plan: admit for observation and IV nitroglycerin.}
\end{quote}

Using LIME, we generated 500 perturbed versions of this text by randomly masking various subsets of tokens. For each, we computed the model’s predicted probability for the ``long stay'' class. Surprisingly, many of these syntactically broken or semantically incoherent variants received high-confidence predictions.

For example, the following perturbed version lacking both grammatical structure and medical interpretability received a predicted probability of 0.778:

\begin{quote}
\texttt{Patient a 65-year-old with type diabetes, , chronic disease. presented with pain. Vitals. Labs: CK-. : for nitroglycerin.}
\end{quote}

Despite losing essential semantic content (e.g., the type of chest pain, presence of dyspnea, and treatment plan), the model remains confident. In total, 86 out of 500 perturbations (17.2\%) exceeded the 0.75 confidence threshold for ``long stay''. A subset of these included only 10 tokens, with some bordering on complete nonsensicality.

This demonstration highlights a fundamental flaw in perturbation-based explainability methods: they often construct examples far outside the data manifold on which the model was trained. Although the model returns a valid prediction for these inputs, the results are misleading from a human interpretability standpoint. In clinical applications, this undermines the goal of generating trustworthy, semantically grounded explanations.

\section{Generalizable Insights about Responsible Application of Machine Learning in Healthcare}

The demonstrations highlight broader lessons for responsible ML in healthcare, showing how common explanation methods behave under realistic clinical text conditions. Six recurring challenges emerge:

\begin{enumerate}
    \item \textbf{Interpretation of feature attributions.} Demonstration~\hyperref[dem1]{1} shows that confidence depends mainly on the top few tokens, while many others carry negligible influence. Presenting long ranked lists risks over-interpreting weak signals. Explanations should better separate dominant drivers from marginal features.

    \item \textbf{Length and complexity of clinical reports.} As seen in Demonstration~\hyperref[dem2]{2}, LIME often highlights trivial content such as stopwords or formatting tokens in lengthy discharge summaries. Such attributions obscure clinically meaningful information in complex inputs.

    \item \textbf{Longer textual units.} Demonstration~\hyperref[dem2]{2} also illustrates that token-level attribution misses multi-word concepts (e.g., ``admit for observation''), instead emphasizing isolated words. Phrase- or entity-level methods are needed for clinically faithful explanations.

    \item \textbf{Off-manifold perturbations.} Demonstration~\hyperref[dem3]{3} shows that perturbed, incoherent notes can still yield confident predictions. Such off-manifold examples distort explanations and threaten trust in clinical deployment.

    \item \textbf{Dilution through averaging.} The deletion analysis in Demonstration~\hyperref[dem1]{1} and the perturbation results in Demonstration~\hyperref[dem3]{3} reveal that averaging across many samples blurs the effect of key tokens and may capture spurious associations, producing diffuse explanations.

    \item \textbf{Instability.} Because LIME relies on random sampling, Demonstration~\hyperref[dem3]{3} indirectly shows that explanations can vary across runs. In healthcare, this instability undermines reproducibility and accountability.
\end{enumerate}

These issues are not isolated artifacts but stem from structural limitations of token-level and perturbation-based explainers. Responsible application in healthcare requires explanation methods that are linguistically coherent, clinically informed, and stable across runs. 

Moreover, similar concerns extend beyond medicine: in legal or scientific NLP, explanations may likewise highlight trivial words, ignore multi-word expressions, or vary under perturbation. Thus, the challenges observed here point to general risks for post-hoc explainability across high-stakes textual domains.

\section{Conclusion and Further Work}
\label{conclusion}

In this study, we examined how post-hoc explanation methods behave when applied to clinical text classification. Through three targeted demonstrations using LIME, we revealed several limitations that can undermine the reliability and interpretability of explanations in healthcare settings.

Our findings showed that explanations often rely on a small number of impactful tokens, while the majority of highlighted features carry minimal influence. We also observed that LIME frequently assigns high attribution to stopwords or function words, which lack standalone interpretive value. Most notably, we demonstrated that syntactically broken or semantically incoherent perturbations can still yield high-confidence predictions, raising serious concerns about explanation validity.

These issues reflect broader structural challenges with token-level and perturbation based approaches in clinical NLP. To address them, future methods should move beyond isolated tokens and incorporate meaningful linguistic or conceptual units such as phrases, medical entities, or clinically grounded abstractions. Additionally, integrating domain knowledge and improving the plausibility of perturbed inputs may help produce more stable and trustworthy explanations.

Ultimately, building clinically useful interpretability tools will require not only technical improvements, but also human centered evaluation, including feedback from medical experts to assess whether explanations align with clinical reasoning and decision support needs. Beyond these directions, future work should also engage with broader responsible ML dimensions, such as ensuring explanation fairness across patient demographics, exploring privacy preserving interpretability techniques, and systematically evaluating potential clinical safety risks that may arise from misleading or unstable explanations.

\section*{Acknowledgments}
Kristian Miok was supported by the EU HE MSC Postdoctoral Fellowship Programme SMASH (No. 101081355).  The work was partially supported by the Slovenian Research and Innovation Agency (ARIS) core research programme P6-0411 and P2-0103, as well as projects L2-50070, GC-0002, and J4-4555. The work was supported by EU through ERA Chair grant no. 101186647 (AI4DH) and cofinancing for research innovation projects in support of green transition and digitalisation (project PoVeJMo, no. C3.K8.IB).

\bibliographystyle{splncs04}
\bibliography{mybibliography}

\end{document}